# On-Board Vision Processing For Small UAVs: Time to Rethink Strategy

Shoaib Ehsan and Klaus D. McDonald-Maier
School of Computer Science & Electronic Engineering,
University of Essex,
Colchester, UK
sehsan@essex.ac.uk, kdm@essex.ac.uk

*Abstract*—The ultimate research goal for unmanned aerial vehicles (UAVs) is to facilitate autonomy of operation. Research in the last decade has highlighted the potential of vision sensing in this regard. Although vital for accomplishment of missions assigned to any type of unmanned aerial vehicles, vision sensing is more critical for small aerial vehicles due to lack of high precision inertial sensors. In addition, uncertainty of GPS signal in indoor and urban environments calls for more reliance on vision sensing for such small vehicles. With off-line processing does not offer an attractive option in terms of autonomy, these vehicles have been challenging platforms to implement vision processing on-board due to their strict payload capacity and power budget. The strict constraints drive the need for new vision processing architectures for small unmanned aerial vehicles. Recent research has shown encouraging results with FPGA based hardware architectures. This paper reviews the bottle necks involved in implementing vision processing on-board, advocates the potential of hardware based solutions to tackle strict constraints of small unmanned aerial vehicles and finally analyzes feasibility of ASICs, Structured ASICs and FPGAs for use on future systems.

## I. Introduction

"Dull, Dirty, Dangerous" [1, 2]-This proverbial statement illustrates very concisely the nature of missions undertaken by unmanned aerial vehicles. Although limited to military applications initially, these vehicles are increasingly viewed as target platform for wide range of civilian applications. The broad application spectrum and aspiration for cost-effective solution have been key factors in development of variants ranging from small vehicles like WASP [3], weighing less than a pound, to Predator-like systems [4] weighing more than 40,000 pounds.

The US Department of Defense describes UAV as "a powered, aerial vehicle that does not carry a human operator, uses aerodynamic forces to provide vehicle lift, can fly autonomously or be piloted remotely, can be expendable or recoverable, and can carry a lethal or non-lethal payload" [5]. Target tracking [6], tactical reconnaissance [7], surveillance [8], search and rescue [9], monitoring of high-tension power lines [10], photogrammetry [11] and remote sensing [11] are some application areas that already employ UAVs.

Irrespective of the specific application, ultimate goal of an unmanned aerial vehicle is autonomy of operation. However, what actually constitutes UAV autonomy is a key question to be asked. According to [5], UAV autonomy is divided into ten levels with every level defining UAV sophistication as shown in Figure 1. Remotely Guided UAVs are at the lowest level of autonomy due to increased human intervention where as Fully Autonomous Swarms are at the highest level. However, it is evident from Figure 1 that most sophisticated UAVs like Predator and Global Hawk can be described at best as semi-autonomous and are yet to achieve full autonomy.

As for humans, the sense of vision is critical for UAVs, not only for accomplishment of assigned missions but also for achieving the autonomy objective. UAVs are generally equipped with still and/or video camera(s) that capture crucial image information. Research in the last decade has proved potential of vision sensing as major contributor to UAV autonomy [12, 13, 14 and 15]. One example is vision based automatic take-off and landing [14] that has reduced human involvement in take-off and landing procedures to minimum, making UAV more autonomous.

In this paper, the focus is on battery-powered small unmanned aerial vehicles (mini and micro) with payload capacity less than 2~3 pounds and flight-time dependent upon system power consumption. Gross weight of such vehicles is also very limited, allowing only light weight, small size batteries with good power to weight ratio to be used on-board. Lithium polymer batteries are quite promising in this regard [16]. With extremely restricted power budget of small UAVs, any computation intensive task carried on-board results in high power consumption leading to small flight-time. AeroVironment's Dragon Eye is one such example which has total weight of 4.5 pounds, payload capacity of 1 pound and can fly for 45 to 60 minutes [5]. The strict payload capacity and small power budget, however, does not change the principal objective - autonomy of operation. Thus, with restricted on-board resources, these battery-powered small unmanned aerial vehicles strive to achieve the levels of autonomy that are still untouched by any other modern day UAV as shown in Figure 1.

## II. Vision Sensing For Small UAVs

As with all UAVs, small unmanned aerial vehicles rely heavily on vision sensing for mission accomplishment. However, for small UAVs, this is even more critical as illustrated by following reasons:

- Due to their strict payload capacity, small unmanned aerial vehicles utilize accelerometers and rate gyros for attitude estimation that weigh few grams but are fairly low grade as compared to their counterparts used on larger aircrafts. These sensors are very sensitive to temperature variations and have high drift rates causing stabilization problems for small UAVs. In [17], it is demonstrated that vision sensing is a practical solution to this stabilization problem.
- Miniaturized dimensions of small UAVs render them suitable for use in urban environments and indoor applications. The uncertainty of GPS (Global Positioning System) signal in such environments leads to navigation problems for these vehicles. Vision based/aided Airborne Simultaneous Localization and Mapping (SLAM) techniques have emerged that allow navigation in absence of GPS signal [18].

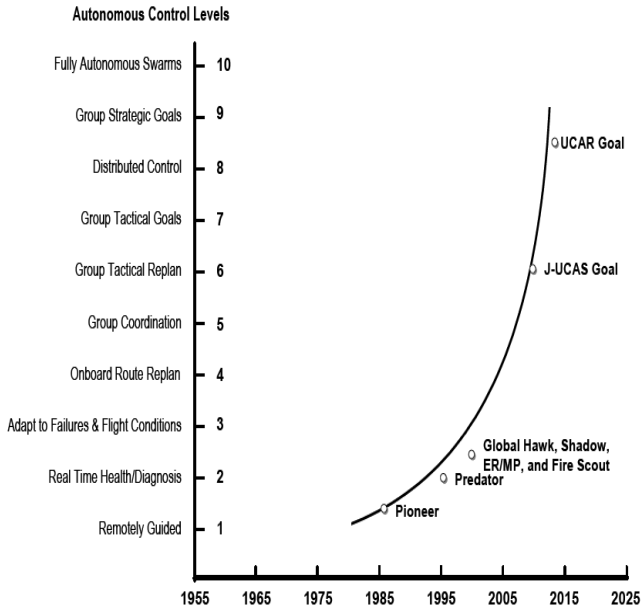

Figure 1. Trend in Unmanned Aerial Vehicle Autonomy [5].

Sense of Vision is, therefore, integral to small unmanned aerial vehicles and has potential of providing viable solutions to problems arising from miniaturized dimensions of such vehicles and the environments in which they typically operate in.

A. *Vision Processing Models for Small UAVs*

Two vision processing models are generally employed for small unmanned aerial vehicles:
- Tele-operation model
- On-board model

In case of Tele-operation model, video and/or still images captured by on-board cameras are compressed using a standard compression technique like JPEG, Wavelet, MPEG, and are transmitted through a wireless link to a ground station. Image/video compression, however, is a computation intensive task and has adverse effect on system power consumption. Moreover, compressed image/video data still requires sufficient bandwidth for wireless transmission that increases pressure on system power resources. Image/video data received on ground station is, therefore, generally noisy and delayed [16]. After carrying out vision processing on a ground computer, tele-commands are then sent back to UAV. Tele-operation is commonly used for small unmanned aerial vehicles but at best can only guarantee near real-time operation. In addition, increased human involvement in Tele-operation is a major drawback, thus, compromising the autonomy objective.

In contrast to Tele-operation, On-board vision processing model encourages vision processing to be carried out on-board as shown in Figure 2. This model ensures real-time operation and autonomy, but is practically realizable only for simple image processing operations due to extremely restricted on-board computational resources. Even for simple image processing operations, keeping system power consumption with in reasonable limits (usually less than a watt) is a non-trivial task. Since autonomy is main driving factor of all UAV related research, this paper focuses on bottlenecks in implementation of vision processing on-board a small UAV, and possible solutions.

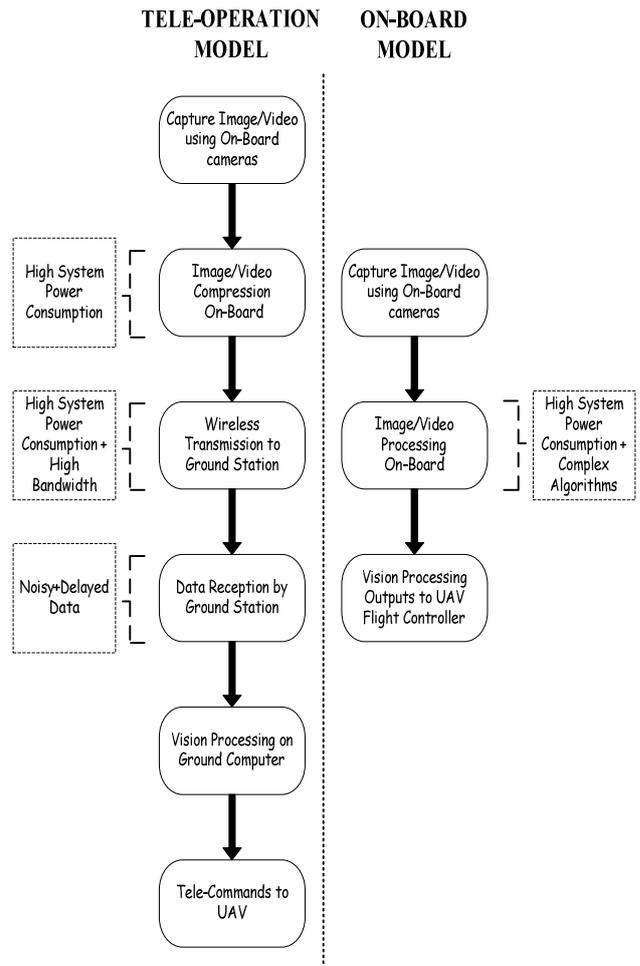

Figure 2. Vision Processing Models for Small UAVs.

*B. Bottlenecks in On-Board Vision Processing*

Image processing and computer vision algorithms are computation intensive and data intensive in nature. Even today, desktop computers with multiple processors running at GHz frequencies, deep memory hierarchies, deep pipelines and huge power budget as compared to small UAVs, are challenged to run low-level computer vision algorithms like feature extraction at rates higher than 2-3 Hz for medium resolution images. COTS high-performance embedded computers that run at relatively lower clock frequencies (usually less than 500 MHz) with moderate pipelines and on-chip memory don't fare any better as their hardware and software architectures are not optimized for vision processing. With a simple, low-power RISC processor on-board with limited computational capability, small UAVs are not suited to process image/video data in real-time. This shows severity of problem at hand. Lack of computer architectures capable of processing image/video data in real-time is major hurdle in vision processing on-board a small UAV.

To make matters worse, strict power constraints demand not only a real-time solution but also a low-power one in order to ensure maximum flight time. A real-time vision processing architecture with high power consumption will decay system battery very quickly that might result in an incomplete mission. Since dynamic power consumption is proportional to clock frequency in case of CMOS, power constraints also limit the maximum clock frequency at which the architecture may operate. This is a major limitation in a sense that one method of enhancing performance is to increase system clock frequency but it is not suitable for use in battery-powered small UAVs. With miniaturized dimensions of this class of vehicles, issues of area and weight are of prime importance as well. A light weight solution with small form factor is what is required. In short, low-power vision processing architectures, operating at low clock frequencies, with small size and light weight (only a few grams), capable of real-time operation, are fundamental to the thought of making small UAVs autonomous vehicles in true sense.

III. GENERAL SOLUTIONS & THEIR VIABILITY ANALYSIS

This section details what can be possible solutions to the problems arising from strict constraints of small UAVs. Generally, there can be four types of solutions to any computation related problem:
- Algorithm based solution
- Software based solution
- Hardware based solution
- Hybrid solution

Theoretically speaking, a pure algorithm based solution is the best possible way forward in small UAVs. Development of new computer vision algorithms that require substantially less expensive computations and extremely reduced data, can pave way for real-time execution on COTS low-power embedded computers. However, it seems a far-fetched idea as well since Computer Vision as a field is still developing despite making substantial advances in the last decade or so and research emphasis is on finding principle solutions, rather than real-time embedded solutions. Vision algorithms focusing on execution speed like Speeded Up Robust Features (SURF) [19] have started to emerge lately but don't ensure real-time performance even on desktop computers. Thus, pure algorithm based approach does not offer a viable solution at this point of time.

A pure software based solution is a good approach to tackle any computational bottleneck but has its limitations due to heavy reliance on computational capabilities of underlying hardware. As a matter of fact, COTS low-power embedded computers have little vision processing capabilities to offer at the moment which actually reduces the probability of a pure software solution for small UAVs.

Recent research has shown that pure hardware based and hybrid solutions have great potential to fit in the limitations of small UAVs. A hybrid solution here means any combination of the three other possible solutions like hardware-software co-design or hardware solution with some algorithmic optimizations to reduce computations. While Parallel hardware structures enable real-time operation, keeping power consumption down to a reasonable limit (usually less than a watt) is still a big challenge. Owing to low development cost and short design time, FPGAs have emerged as a promising solution with ability to satisfy strict size, weight and power constraints achieving real-time performance for small unmanned aerial vehicles. The following examples demonstrate the encouraging results obtained with FPGA based solutions. In [20], a vision based drift stabilization control system is implemented on a custom, low-power FPGA board employing Virtex-4 for a quad-rotor micro unmanned aerial vehicle. A hardware-software co-design strategy is used with a Harris feature detector, Template matching, RGB to HSV conversion and color segmentation algorithms implemented in hardware while RANSAC and Kalman filter run on PowerPC inside Virtex-4. However, no detail about power consumption of this design is provided. In [21], a vision system for precision MAV targeted landing is implemented on a Virtex-4 FPGA using a hardware-software co-design approach. Again, no detail is given about power consumption of design.

IV. TIME TO RETHINK STRATEGY

Today, FPGAs are not the only available hardware based solution but interestingly, other options have not generally been considered for small UAVs due to low development cost and short design time factors of FPGA. Clearly, FPGA has shown good potential for vision processing on small UAVs and better results may be obtained in future; however, it is time to explore other prospective solutions as well due to following factors:
- The desire to achieve the autonomy objective drives the need to implement more and more complex vision processing on-board a small UAV that might one day lead to an integrated vision based Guidance, Navigation and Control system on a single chip. As complexity of on-board vision processing increases, so will increase pressure on system's power consumption. Although, FPGAs have so far proved a good choice for low-level vision processing algorithms like feature extraction, there is still a question mark regarding its capability to implement more complex vision processing in future with strict power constraints.

- According to Moore's law [22], transistor densities double every two years. However, with decreasing transistor geometries, static power consumption in deep sub-micron technologies is coming up as a major challenge for Silicon industry. In [23], it is predicted that static power consumption will become a significant contributor to overall power consumption for deep sub-micron technologies due to leakage currents as shown in Figure 3. All hardware based solutions will face the challenge posed by static power consumption; however, FPGAs appear to be the most susceptible as they have currently a higher power consumption relative to other competing solutions. The static power factor might hinder implementation of complex vision processing with low power consumption on FPGAs.

With the above given factors in mind, there is a strong motivation to explore existing hardware based solutions other than FPGA that might be used in future on small unmanned aerial vehicles. ASIC/SoC and Structured ASIC are two solutions that appear to have potential to handle future power consumption challenges much better than FPGAs but have not traditionally been considered for small UAVs. Structured ASICs, manufactured by big companies like Altera, AMI, ChipX, eASIC, Faraday, Fujitsu and NEC, has emerged as a mid-way between ASIC and FPGA [24]. Therefore, it is worth while to carry out a feasibility analysis of ASIC, Structured ASIC and FPGA focusing on strengths and weaknesses of all three competing solutions for future small UAVs.

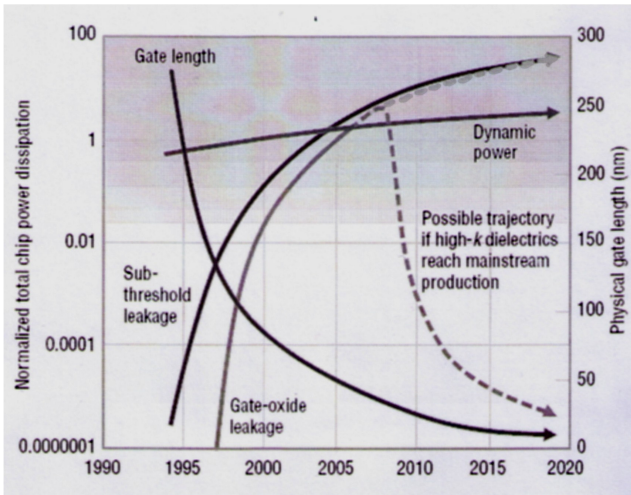

Figure 3. Total chip dynamic and static power dissipation trends based on the International Technology Roadmap for Semiconductors [23].

## V. HUNTING FOR FUTURE SOLUTION

This comparison is based on physical and computational constraints of future unmanned aerial vehicles. Specific attributes of competing solutions that play an important role in meeting these strict constraints are evaluated to identify pros and cons of each solution.

### A. Development Cost

It is considered one of the major factors in making FPGA a solution of choice so far as small UAVs are meant to be low cost. A high end FPGA board like Virtex-5 board costs as low as $ 995 only [25]. On the other hand, ASIC has high NRE costs and according to [26], large upfront cost is estimated to be around $25M. According to [27], cost of ASIC is $20M or more. Structured ASIC is claimed to have only 25% of development cost of ASIC [28]. As an example, for the three competing solutions, a comparison of development cost for typical 1-million gate design in 0.13 µm technology is given in Table I suggesting that FPGA is an obvious winner in this regard [29].

### B. Production Cost

Small UAVs are gaining more and more popularity due to their broad application spectrum and are expected to increase in number in future. Thus, production cost in a volume application is an important factor to consider while searching for a long-term solution. Of the three competing solutions, ASIC has lowest per unit cost in high production runs and FPGA has the highest [28]. Structured ASIC has low production cost for low-medium and medium-high production runs. According to [30], Structured ASICs are feasible for production volumes of less than 100,000. Production cost of Structured ASIC is projected to be only 10% of that of FPGA [28]. This is illustrated by a comparison of unit cost for typical 1-million gate design in 0.13 µm technology as given in Table II for ASIC, Structured ASIC and FPGA indicating dominance of ASIC [29]. With demand of small UAVs expected to increase in future, ASIC would therefore, become an increasingly cost-effective solution.

### C. Transistor Density

Keeping in mind complexity of vision systems on future UAVs, transistor density is a key factor not only in providing ability to accommodate big designs but also in satisfying strict size constraints. ASIC is outstanding in this regard with very high transistor densities that will soon touch One billion transistors per square cm mark as shown in Figure 4 [31]. Structured ASIC and FPGA are far behind with a factor of around 3:1 and 100:1 respectively as compared to ASIC [28].

### D. Design Area

Design area is an important consideration when area constraint is involved as in the case of small UAVs. ASIC turns out to be the best in terms of design area efficiency. According to [32], designs that are implemented in FPGA using LUTs (Look-Up Tables) and flip-flops only are estimated to be 40 times larger than ASIC but a substantial decrease in area can be achieved by using hard multipliers and dedicated memories. In contrast, the design area for Structured ASIC is claimed to be two times larger than ASIC implementation [33].

TABLE I. COMPARISON OF DEVELOPMENT COST FOR A TYPICAL 1-MILLION GATE DESIGN IN 0.13 MICRON TECHNOLOGY [29]

| Development Cost | FPGA | Structured ASIC | Cell-based ASIC |
|---|---|---|---|
| Total Design Cost | $165 K | $500 K | $5.5 M |
| Vendor NRE | None | $100 K ~ $200 K | $1 M ~ $3 M |
| Cost of EDA Tools | $30 K | $120 K ~ $250 K | > $300 K |
| Man Power | 1 to 2 | 2 to 3 | 5 to 7 |
| Price per Chip | $200 ~ $1 K | $30 ~ $150 | $30 |

TABLE II. COMPARISON OF PRODUCTION COST FOR A TYPICAL 1-MILLION GATE DESIGN IN 0.13 MICRON TECHNOLOGY [29]

| Production Cost | FPGA | Structured ASIC | Cell-based ASIC |
|---|---|---|---|
| Unit Cost (Qty 1K) | $1000 | $500 ~ $650 | $55 K |
| Unit Cost (Qty 5K) | $220 | $110 ~ $150 | $1.1 K |
| Unit Cost (Qty 500K) | $40 | $21 | $11 |

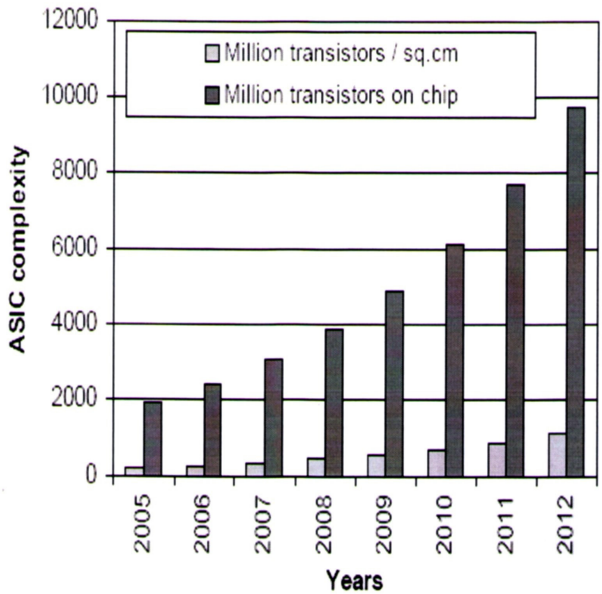

Figure 4. ASIC Design Complexity [31].

*E. Power Consumption*

Although still a vital issue, power consumption may become decisive factor in establishing a vision processing solution for small UAVs in future due to increasing complexity of vision systems and significant static power consumption in deep sub-micron technologies as mentioned above. FPGAs typically dissipate 10 to 15 times more power than equivalent ASIC implementation [28]. On the other hand, Structured ASICs are estimated to consume 2 to 3 times more power than their ASIC counterparts [28]. According to [34], FPGA will never challenge ASIC for an application where low-power is a serious concern. With static power consumption predicted to play considerable role in deep sub-micron technologies, FPGAs will be affected more in terms of power consumption as compared to the other competing solutions. Xilinx, a leading FPGA vendor, has turned to 'Triple oxide Process Technology' and a new 6-input Look-up table (LUT6) architecture to minimize static power consumption in Virtex-5 devices at 65 nm. Xilinx claims that static power consumption of Virtex-5 devices is comparable to older generation Virtex-4 devices (at 90 nm), however, no detail is provided regarding percentage difference in static power consumption between two device generations [35]. Altera, another major FPGA vendor, has come up with 'Programmable Power Technology' to tackle static power consumption problem. As claimed by Altera, Stratix IV at 40 nm consumes 37% less static power with 'Programmable Power Technology' than expected as shown in Figure 5 [36]. Irrespective of the level of authenticity of this claim, an important point to consider is the fact that there is an increase in static power consumption even with the use of 'Programmable Power Technology' and this will come into play more and more as transistor geometries continue to decrease. If the current trend continues for the next few years, FPGAs might struggle as a low power solution. In addition, designs implemented in FPGAs with LUTs and flip-flops only, consume 12 times more dynamic power on average than equivalent ASIC implementations although some improvement is possible by using hard multipliers and dedicated memories [32]. Thus, ASICs and Structured ASICs have a big edge over FPGAs in terms of power consumption and might get preference over FPGAs for use in small UAVs in near future.

*F. Design Time*

Design time plays an important role in achieving fast time to market objective. Of the three alternatives being considered, FPGAs have the shortest design cycle time due to pre-fabricated nature. Due to this very fact, it is used as rapid prototyping platform for ASICs as well. Structured ASICs also have this pre-fabrication advantage but to a lesser extent in a sense that they still require limited number of metallization layers to complete them. ASICs typically require design time of around 2 years owing to the physical design stages involved and followed by manufacturing [27]. FPGA, therefore, seems most suitable when considering design time factor for future systems.

*G. Turn Around Time*

FPGA is hard to compete with in this regard. FPGAs usually require turn around time of 1 to 4 weeks [37]. The turn around time for Structured ASICs is about 2 months [29] whereas ASICs require 2 to 5 months [37]. For future systems, long turn around times can make it hard to achieve fast time to market objective; therefore, use of FPGAs can be a solution to counter this problem.

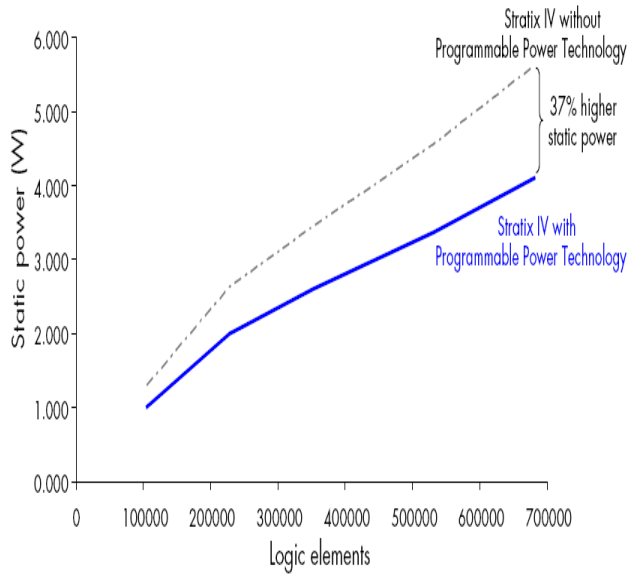

Figure 5. Static Power Consumption in Stratix-IV FPGA [36].

*H. Speed*

Performance of design in terms of speed (clock frequency) can be a significant factor in achieving real-time on-board processing for small UAVs with more complex vision systems in future. FPGAs are capable of only 10% to 20% of performance of ASIC. Structured ASICs perform better achieving 70% to 80% of performance of ASIC [28]. Thus, ASIC turns out to be the solution of choice in terms of performance.

## VI. CONCLUSIONS

On-board vision processing is fundamental to the concept of autonomous unmanned aerial vehicles. Small UAVs strive for autonomy with extremely restricted on-board resources and it is a big challenge for them to carry out a computation intensive task like vision processing. FPGAs have emerged as a low development cost solution that has potential of satisfying strict physical and computational constraints of small UAVs. This paper has tried to explore other potential existing hardware based solutions that have not traditionally been considered for small UAVs. Strengths and shortcomings of ASICs, Structured ASICs and FPGAs were assessed that indicate the feasibility of ASICs and Structured ASICs for use on small UAVs in future.

It is evident from the above comparison that ASICs and Structured ASICs dominate FPGAs overwhelmingly in terms of power consumption, transistor density, design area and speed-all expected to be problem areas for FPGAs in future. If popularity and demand of small UAVs continue to rise as expected, ASICs and Structured ASICs would become increasingly economical solutions owing to their low production cost as compared to FPGAs. In short, ASICs and Structured ASICs are well-equipped to tackle future challenges for small UAV and its time to re-think strategy regarding on-board vision processing solutions. Since ASIC based designs are typically portable to FPGA but vice-versa is not true, the best possible way forward in the near term will be to design fully synthesizable vision processing systems for ASIC implementation and continue using FPGAs as rapid prototyping platforms. This will lead to a smooth transition from FPGA to ASIC or Structured ASIC in case of failure of FPGAs to satisfy strict constraints of small UAVs in future.

ACKNOWLEDGMENT

The authors would like to thank Dr. Adrian F. Clark for his helpful suggestions and discussions.